\documentclass[wcp]{jmlr}
\usepackage{graphicx}
\usepackage{booktabs, array}
\usepackage{multirow}
\usepackage{siunitx}
\usepackage{float}
\usepackage{enumitem}
\usepackage{hyperref}
\usepackage{amsmath,bm}
\usepackage[normalem]{ulem}
\useunder{\uline}{\ul}{}
\makeatletter
\let\Ginclude@graphics\@org@Ginclude@graphics
\renewcommand*{\thanks}[1]{%
  \footnotemark
  \protected@xdef\@thanks{\@thanks
    \protect\footnotetext[\arabic{footnote}]{#1}}%
}
\makeatother

\title[Examining the Effect of Pre-training on Time Series Classification]{Examining the Effect of Pre-training Followed by Fine-tuning on Time Series Classification}

 \author{\Name{Jiashu Pu} \Email{pujiashu@corp.netease.com}\\
  \Name{Shiwei Zhao} \Email{zhaoshiwei@corp.netease.com}\\
  \addr Fuxi AI Lab, Netease Inc.
  \AND
 \Name{Ling Cheng} \Email{lingcheng.2020@phdcs.smu.edu.sg}\\
\addr Singapore Management University
\AND
  \Name{Yongzhu Chang} \Email{changyongzhu@corp.netease.com}\\
  \Name{Runze Wu} \Email{wurunze1@corp.netease.com}\\
  \Name{Tangjie Lv} \Email{hzlvtangjie@corp.netease.com}\\
  \Name{Rongsheng Zhang}\thanks{Corresponding Author} \Email{zhangrongsheng@corp.netease.com}\\
  \addr Fuxi AI Lab, Netease Inc.
  }
 
\begin{document}
\maketitle

\begin{abstract}
Although the pre-training followed by fine-tuning paradigm is used extensively in many fields, there is still some controversy surrounding the impact of pre-training on the fine-tuning process. Currently, experimental findings based on text and image data lack consensus. To delve deeper into the unsupervised pre-training followed by fine-tuning paradigm, we have extended previous research to a new modality: time series. In this study, we conducted a thorough examination of 150 classification datasets derived from the Univariate Time Series (UTS) and Multivariate Time Series (MTS) benchmarks. Our analysis reveals several key conclusions. (i) Pre-training can only help improve the optimization process for models that fit the data poorly, rather than those that fit the data well. (ii) Pre-training does not exhibit the effect of regularization when given sufficient training time. (iii) Pre-training can only speed up convergence if the model has sufficient ability to fit the data. (iv) Adding more pre-training data does not improve generalization, but it can strengthen the advantage of pre-training on the original data volume, such as faster convergence. (v) While both the pre-training task and the model structure determine the effectiveness of the paradigm on a given dataset, the model structure plays a more significant role. 


\end{abstract}





\section{Introduction}
\emph{The pre-training then fine-tuning paradigm} continues to shine in the Natural Language Processing field, owing to the immense data and extra-large model sizes, with ultra-large models dominating the SuperGlue Benchmark~\cite{wang2019superglue}. Recently, the masked autoencoding pre-training scheme has also demonstrated its viability in Computer Vision~\cite{he2022masked}. Deep Learning seems to be moving towards a grand unification of pre-training. Nonetheless, there is still a lack of consensus on how exactly this paradigm manifests itself. While earlier work~\cite{bengio2006greedy} has shown that unsupervised pre-training helps optimization, the later milestone research~\cite{erhan2010does} argues that unsupervised pre-training improves generalization by acting as a form of regularization. In contrast, a recent study ~\cite{he2019rethinking} claims that supervised pre-training only improves convergence speed without benefiting generalization. Another study~\cite{abnar2021exploring} showcases a scenario where the downstream performance is at odds with the supervised pre-training accuracy. Other works suggest pre-training offers more obvious advantages of combating label noise~\cite{hendrycks2019using}, alleviating catastrophic forgetting~\cite{mehta2021empirical}, and dealing with imbalanced datasets~\cite{liu2021self}.

The divergent conclusions drawn from prior research have prompted us to investigate this issue using a different approach. Our curiosity lies in determining which findings demonstrate cross-modal consistency. Time series possess two distinct properties that complicate their analysis. Firstly, the features of time series data vary significantly across different domains, posing a challenge to domain transfer~\cite{eldeletime}. Secondly, within the same domain, the distribution of time series data can shift over time~\cite{tonekaboni2020unsupervised}, rendering the use of long-standing data less effective. It is likely due to these factors that \emph{the pre-training and then fine-tuning paradigm} has yet to prosper in the field of time series.  Nonetheless, we believe that further investigation into this direction is worthwhile. Specifically, we focus on the widely concerned problem of Time Series Classification~(TSC) to understand how this paradigm operates. Notwithstanding its high accuracy on the UCR benchmark~\cite{dau2019ucr}, the current best ensemble model of Time Series Classification, HIVE-COTE 2.0~\cite{middlehurst2021hive}, suffers from slow training speed and challenging deployment. Moreover, given the complexity of time series data~\cite{tonekaboni2020unsupervised}, it is difficult for non-specialists to directly annotate the raw time series data~\cite{eldeletime,zerveas2021transformer}, even as time series data relevant to our daily lives continues to accumulate at an unprecedented rate~\footnote{\url{https://www.forbes.com/sites/forbescommunicationscouncil/2022/06/16/the-ubiquity-of-time-series-data-isnt-coming-its-already-here}}. We posit that \emph{the unsupervised pre-training and fine-tuning paradigm} holds promise for the future of Time Series Classification. Considering the characteristics of time series data, we design a pre-training setup with in-domain datasets, conducting experiments on 150 datasets with three model structures and five pre-training tasks.
\textbf{Our study makes a threefold contribution}. Firstly, we verify the feasibility of the paradigm of unsupervised pre-training followed by fine-tuning on TSC. Secondly, we re-validate some existing conclusions concerning the impact of unsupervised pre-training on fine-tuning for time series data. Additionally, we provide novel insights into which factors - the pre-training task or model structure - are more critical in enhancing the efficacy of pre-training on fine-tuning. Lastly, we attempt to find correlates of successful pre-training.
\textbf{Our key findings are as follows:}\\
- Pre-training can enhance optimization for under-fitted models of few parameters (Consistent with~\cite{bengio2006greedy} under constraints), but it does not improve optimization for  for models that already have adequate ability to fit the data. \\
- Given sufficient training time, pre-training does not significantly improve a model's generalization ability (Consistent with~\cite{he2019rethinking}), i.e., it has no regularization effect (Contradict with~\cite{erhan2010does}).\\
- Pre-training can accelerate the convergence speed (Consistent with~\cite{he2019rethinking}), but only if the model has the capability to fit the data well. \\
- Increasing the amount of pre-training data does not aid generalization (Contradict with~\cite{paine2014analysis}), but it can strengthen the existing advantages of pre-training for the original data sizes.\\
- The effectiveness of \emph{the pre-training followed by fine-tuning paradigm} on a given dataset is determined by both the model structure and the pre-training task, with the model structure being more crucial.

\section{Related Work}
Unsupervised pre-training is gaining more and more attention in the field of time series. Feature-wise, most of the work is based on temporal features, considering both inter-sample dissimilarity and intra-temporal resemblance to design pre-training tasks around contrastive learning~\cite{yue2022ts2vec,eldeletime,tonekaboni2020unsupervised}. The time-frequency interchangeability motivates other work to exploit the frequency domain to obtain proper representations of the seasonal trend~\cite{woo2021cost,zhang2022self}. However, most of the work above uses linear probing to verify the effectiveness of representations on tasks such as classification~\cite{eldeletime,franceschi2019unsupervised}, forecasting~\cite{yue2022ts2vec}, regression~\cite{zerveas2021transformer}, and anomaly detection~\cite{yue2022ts2vec}.In the field of time series, no work has investigated how unsupervised pre-training affects fine-tuning.

In other fields such as Natural Language Processing and Computer Vision, it is still controversial how unsupervised pre-training works on fine-tuning. Some work suggests that unsupervised pre-training is beneficial for optimization~\cite{hao2019visualizing,neyshabur2020being}, while other work argued that unsupervised pre-training only serves the function of regularization~\cite{erhan2010does}. There is also evidence suggesting that unsupervised pre-training can only speed up convergence~\cite{he2019rethinking}. In the face of adversarial samples, imbalanced datasets~\cite{liu2021self}, and label corruption, some work also argues that pre-training has a great advantage and can improve the model's uncertainty estimates as well~\cite{hendrycks2019using}. In addition, in the life-long learning scenario, one work suggests that pre-training can alleviate the effects of catastrophic forgetting~\cite{mehta2021empirical}. Our work extends the above efforts by validating some existing ideas in the field of time series while doing new research on finding potential correlates of effective pre-training, the inductive bias of pre-training tasks, etc.

\section{Problem Formulation}
Our paper focuses on the validity of the \emph{paradigm of unsupervised pre-training followed by fine-tuning} for the Time Series Classification task. To this end, we verify its effectiveness on 150 datasets.
In the following paragraphs, we first introduce the notations of the time series data and the encoder, then we present how experiments are designed and what issues we specifically analyze.

\noindent \textbf{Time Series Data} We define a group of continuous variables changing over time as $s$, where $s \in \mathbb{R}^{d}$, with $d$ corresponding to the number of variables or, alternatively, the dimension of the features. Based on this, we define a time series of length $\ell$ as $S=(s_{1}, \ldots, s_{\ell})$. When $d$ is $1$, the time series $S$ belongs to the UTS, while when $d$ is greater than $1$, the time series $S$ belongs to the MTS.

\noindent \textbf{Time Series Encoder and Classification Head}
We define the time series encoder as $f_{\theta_1}: S\rightarrow E$, where $E$ is the output of the encoder's last layer. The output $E = (e_{1}, \ldots, e_{\ell})$, where $e_{i} \in \mathbb{R}^{h}$ and $h$ is the hidden dimension of the output layer. To adapt to different classification datasets, we introduce the classification head $g_{\theta_2}: E\rightarrow \{1, \ldots, C\}$ such that $(g_{\theta_2} \circ f_{\theta_1}): S\rightarrow y \in \{1, \ldots, C\}$, where $C$ is the total number of categories and $y$ denotes the categorical output. More specifically, the classification head $g_{\theta_2}$ consists of a one-dimensional convolutional layer followed by a Multiple Layer Perceptron (MLP). By considering the length $\ell$ of the sequence as the number of channels, we apply the one-dimensional convolutional layer to transform the encoder's output tensor $\mathrm{T_{E}} \in \mathbb{R}^{\ell \times h}$ into a length uncorrelated vector $\boldsymbol{v}_{E} \in \mathbb{R}^{h}$. Using the vector $\boldsymbol{v}_{E} \in \mathbb{R}^{h}$ as the input, the MLP layer produces the output as the predicted class label $y$.

\noindent \textbf{ Research Problems and Experimental Procedure}
We chose three sequence encoder structures and five pre-training tasks for our study, with a pre-training task expressed as $\mathcal{PT}$. We denote an unlabelled pre-training dataset as $\mathcal{D}_{pre}$, where $\mathcal{D}_{pre}=(S_1, \ldots, S_{|D_{pre}|})$. Similarly, we define the training set and test set of a classification dataset as $\mathcal{D}_{train}$ and $\mathcal{D}_{test}$. 
There are $150$ experimented datasets in total, for each of them, we iterate through all combinations of different model structures and pre-training tasks. For each combination, we first pre-train $f_{\theta_1}$ on $\mathcal{D}_{pre}$ with $\mathcal{PT}$ and obtain a specific set of pre-trained parameters $\theta_{pre}$, after which the encoder $f_{\theta_1}$ initialized with $\theta_{pre}$ and $g_{\theta_2}$ are simultaneously fine-tuned on $\mathcal{D}_{train}$ (The domain of $\mathcal{D}_{pre}$ is the same as $\mathcal{D}_{train}$, and we set $\mathcal{D}_{train} \subseteq \mathcal{D}_{pre}$ for every dataset, the reason for which is described in Section \ref{sec: Pre-training_detail}). Based on the results of all test sets and the recorded training processes, we perform a significance analysis on whether pre-training is advantageous over random initialization in various aspects---including optimization, generalization, convergence speed, etc. We verify whether the role of pre-training changes when the model is significantly under-fitted, and whether adding additional pre-training data is beneficial. In addition, we also analyze the results to answer a question --- which aspect has more influence on the fine-tuning results given an arbitrary dataset, the model structure, or the pre-training task?

\section{Research Subjects}
\subsection{Time Series Encoder}
While ensemble models continue to set new records on the UCR benchmark~\cite{middlehurst2021hive}, we select 
three classical model structures of moderate complexity as the time series encoder (the number of parameters is around 500,000). There are two reasons.
First, models designed for time series need to be scalable and efficient because signals are often long and have a high dimension~\cite{tonekaboni2020unsupervised,franceschi2019unsupervised}. Second, under limited computational resources, we chose to do experiments on as many datasets as possible to make conclusions informative and credible. It is worth noting that the model structures presented below all model the time series in both directions.

\noindent \textbf{LSTM}: Previous work~\cite{sagheer2019unsupervised} shows the usefulness of unsupervised pre-training of LSTM-based autoencoder for MTS prediction tasks. Here we have simplified the model by constructing a vanilla bidirectional LSTM of two layers.

\noindent \textbf{Dilated Convolutional Neural Network (D.Conv)}: The Convolutional Neural Network performs well on time series forecasting~\cite{yue2022ts2vec} and demonstrates its strengths of representing learning on the UTS and the MTS datasets~\cite{franceschi2019unsupervised}. \emph{D.Conv} consists of layers of dilated convolutions. Compared to the previous work \cite{franceschi2019unsupervised}, we do not use causal convolutions, so as to incorporate information from both before and after time step $i$ when conducting convolution operations. 

\noindent \textbf{Time-Series Transformer (TsTransformer)}: The Time-Series Transformer has proved a success in representing the MTS data type~\cite{zerveas2021transformer}. It has the same structure as the original transformer encoder~\cite{vaswani2017attention}, except that it replaces the Layer Normalization layer with the Batch Normalization layer and the embedding layer with the linear projection layer.

\subsection{Pre-training Task}
Representation learning of time series is becoming more and more sought-after in recent years. We select some recently proposed unsupervised pre-training tasks of relative efficacy, while also taking into account diversity. 
\begin{itemize}[leftmargin=0pt,label={}]
  \item \textbf{(Baseline) Random-Cls~\cite{maennel2020neural}}: \emph{Radom-Cls} is the abbreviation of training with Random Class Labels. Because one paper claims that pre-training the model with random labels can effectively improve the convergence speed when fine-tuning~\cite{maennel2020neural}, we choose it as the baseline of the pre-training task.
  
  \item \textbf{Ts2Vec~\cite{yue2022ts2vec}}: Ts2vec is a contrastive learning-based approach. It highlights the construction of positive and negative samples by taking into account both sample and temporal differences. The other strength of Ts2Vec lies in modeling time series at hierarchical levels of granularity. This method has been evaluated in great detail on the UTS and MTS datasets, but it only validates its effectiveness in the linear-probing setting.
  
  \item \textbf{Ts-Tcc~\cite{eldeletime}}: Ts-Tcc is another contrastive learning framework, integrating two contrasting modules --- temporal and contextual. It starts with generating a strong and a weak view from the time series $S$ via augmentation. During temporal contrasting, the model predicts future series ($S_{>t}$) of one view from another ($S_{\leq t}^{\prime}$), where $S^{\prime}$ denotes another view of $S$ and $t$ is the time step. While the contextual contrasting is to minimize the distance between $f_{C}(S_{\leq t})$ and $f_{C}(S_{\leq t}^{\prime})$, where $f_{C}$ summarize the context till $t$, at the same time, maximize the distance between $f_{C}(S_{\leq t})$ and the context of other augmented views of different instances.
  
  \item \textbf{Mvts~\cite{zerveas2021transformer}}: The abbreviation Mvts comes from the tile `A Transformer-based Framework for \textbf{M}ulti\textbf{v}ariate \textbf{T}ime \textbf{S}eries Representation Learning'. Given a sample of time series $S$, Mvts obtains corrupted inputs by zeroing. For instance, we can mask (zero) every variable independently (for the MTS data type) or mask a subseries of $S$. The pre-training task is to impute these zeroed inputs, which is analogous to the Masked Language Modeling~\cite{devlin2019bert} in NLP. The loss function of Mvts is the Mean Squared Error.

  \item \textbf{Srlt~\cite{franceschi2019unsupervised}}: The abbreviation Srlt comes from the paper title --- Unsupervised \textbf{S}calable \textbf{R}epresentation \textbf{L}earning for Multivariate \textbf{T}ime Series. Inspired by how word2vec~\cite{mikolov2013efficient} is trained, this method proposes a novel triplet loss for modeling the time series data. Roughly speaking, given a time series $S$, its subseries of varying lengths are selected as positive samples, while other subseries of some arbitrary time series are selected as negative samples.
\end{itemize}

\subsection{Given a dataset, how to weigh the picking of the model structure and the design of the pre-training task?}
The pre-training tasks for time series are built on some implicit or explicit assumptions. Ts-Tcc, Ts2Vec, and Mvts all assume that time series are contextually related and predictable, and Ts-Tcc assumes that time series have stretch and shuffle invariance properties. Therefore, we believe that different pre-training tasks have their specific applicability scenarios and limitations. Likewise, different model structures also have different inductive biases~\cite{tuli2021convolutional}. The model structure determines the parameter space, while the pre-training process determines the starting point of fine-tuning. Given an arbitrary dataset, which one is more critical for successful fine-tuning? The pre-training task or the model structure?


To reveal potential correlates, we calculate the Spearman's rank correlation coefficient~\cite{schober2018correlation} between different variables and the relative test set accuracy gaps, which equals the test-set accuracy of the pre-trained model $f_{\theta_{pre}}$ minus the accuracy of randomly initialized model $f_{\theta_1}$. We take sequence length as an example. We denote $M$ as the total number of datasets and $b_i$ as the average sequence length of $i$-th data set. We have the variable of sequence length $B$, where $B=(b_1, \ldots, b_M)$. The other variable of the accuracy gap $A$ can be defined similarly. If we calculate the correlation coefficient between $A$ and $B$ as $0.55$, with the $p$-value being $0.03$ ($<0.05$). We can conclude that $A$ is positively correlated with $B$ and the result is significant enough.

\subsection{Can we determine in advance whether we need to pre-train and when can we trust the pre-trained parameters?}
In the practical application of time series data, we may have abundant unlabeled data, while often not having sufficient annotated data~\cite{eldeletime}. When the real demand is for fast iteration or instant migration to new scenarios, it becomes critical to understand in advance whether $\theta_{pre}$ has the potential to fit the data of the downstream task and generalize beyond. To this end, we collect potential correlates from three perspectives---the data, the pre-training process, and the model parameters, hoping to uncover some indications for effective pre-training. The correlation factors include: the length of time series, the size of the pre-training data, $\ell_2$ norms~\cite{neyshabur2017exploring}, $\ell_2$-path norm~\cite{jiang2019fantastic}, the sharpness value~\cite{mehta2021empirical,hao2019visualizing}, the convergence state of pre-training and the $\ell_2$ distance $\theta_{pre}$ traveled from the initial point. The $\ell_2$ norms, the $\ell_2$-path norm, and the sharpness value are three generalization measures validated by previous work. Recent work~\cite{gouk2020distance,mao2020survey} suggest that traveled distance is associated with generalization performance.

To reveal potential correlates, we calculate the Spearman's rank correlation coefficient~\cite{schober2018correlation} between different variables and the relative test set accuracy gaps, which equals the test-set accuracy of the pre-trained model $f_{\theta_{pre}}$ minus the accuracy of randomly initialized model $f_\theta$. We take sequence length as an example. We denote $M$ as the total number of datasets and $b_i$ as the average sequence length of $i$-th data set. We have the variable of sequence length $B$, where $B=(b_1, \ldots, b_M)$. The other variable of the accuracy gap $A$ can be defined similarly. If we calculate the correlation coefficient between $A$ and $B$ as $0.55$, with the $p$-value being $0.03$ ($<0.05$). We can conclude that $A$ is positively correlated with $B$ and the result is significant enough.

\section{Datasets and Experimental settings}
We use a total of 150 datasets, 125 of which are from the UTS data type and 25 from the MTS data type. Due to hardware constraints, we discard six datasets that contain excessively long and highly dimensional sequences. In most of the datasets, the sequence lengths are equal, while in other few datasets where the sequences are not equal, we pad the sequences to the maximum length with zeros. We present basic statistics of the UTS and MTS datasets in Table~\ref{tab:dataset_statistical}.
\begin{table*}[]
\centering
\setlength{\tabcolsep}{14pt}
\begin{tabular}{@{}ccccc@{}}
\toprule
 & Num.Class & Num.Sample & Series.Length & Feature.Dim \\ \midrule
UTS & 9/2/60 & 473/16/8926 & 794/15/13167 & 1/1/1 \\
MTS & 9/2/39 & 1866/12/25000 & 1159/8/17984 & 99/2/1345 \\ \bottomrule
\end{tabular}
\caption{Basic statistics of the UTS and the MTS datasets. We present the average/min/max value for each aspect. \textbf{Feature.Dim} means feature dimension.}
\label{tab:dataset_statistical}
\end{table*}

Due to the complexity, diversity, and timeliness of time series data~\cite{zhang2022self}, collecting in-domain data depends upon strong expert knowledge~\cite{eldeletime}. Thus it is impractical to collect a large amount of unlabeled data for each dataset, to simplify the experimental setup, we set $\mathcal{D}_{train}$ as $\mathcal{D}_{pre}$ in our main experiments, which is a setup focusing on the low-resource scenario --- an important topic in today's machine learning community~\cite{rotman2019deep}. We divide each $\mathcal{D}_{pre}$ into a training set $\mathcal{D}_{pre}^{train}$ and a validation set $\mathcal{D}_{pre}^{val}$ ($10\%$ of $\mathcal{D}_{pre}$). To avoid over-fitting, we retain the parameter with the lowest validation loss as $\theta_{pre}$. We set the learning rate, batch size, and epoch of pre-training to $1e^{-4}$, $32$, and $100$ respectively. Some pre-training works mention the sensitivity of their methods to hyperparameters~\cite{yue2022ts2vec,zerveas2021transformer}, so we randomly select $30$ UTS datasets, perform a gird search of hyperparameters for each pre-training task and pick the best ones.

For fine-tuning, we split $10\%$ of $\mathcal{D}_{train}$ as the validation set, on which we trace the model with the highest accuracy. Learning rates of $f_{\theta_1}$ and $g_{\theta_2}$ are both set to $1e^{-3}$, and batch size is set to $32$. We apply gradient clipping and set the clip value to $4.0$. We train encoders and classifiers for $200$ epochs to ensure convergence, with the cross-entropy loss. To evaluate our model, we use the official split~\cite{dau2019ucr} of training and test data for each dataset. We train the model five times with different random seeds (control weight initialization) for a given set of model structures and pre-training tasks. The final test result for each dataset is the average of the five runs.
The pre-training and supervised training process share the same optimizer of Adam~\cite{kingma2014adam}. For other detailed configuration of the model structure and hyperparameters please refer to code repository.
\label{sec: Pre-training_detail}

\section{Experimental Results and Analysis}
The Wilcoxon signed rank test~\cite{rey2011wilcoxon} is performed between the results of each pre-trained model and the results of the train-from-scratch counterpart. In Table~\ref{tab:main_result} and Table~\ref{tab:under_fit_LSTM_cnn}, the symbol $\dag$ indicate $p$-value $<0.05$, and the down arrow $\downarrow$ indicates that the pre-trained result is significantly worse ($p<0.05$) than the non-pre-trained one.

\subsection{Effect of pre-training}

\begin{table*}[]
\centering
\setlength{\tabcolsep}{6.3pt}
\begin{tabular}{@{}llllllll@{}}
\toprule
 & \textbf{} & \textbf{Not.P} & \textbf{R.Cls} & \textbf{Ts2Vec} & \textbf{Ts-Tcc} & \textbf{Mvts} & \textbf{Srlt} \\ \midrule
 & \multicolumn{7}{c}{\textbf{Training Loss (Min)}} \\ \midrule
 & LSTM & 0.045 & \textbf{0.043} & {\ul 0.044} & 0.046 & {\ul 0.044} & {\ul 0.044} \\
 & TsTransformer & 0.389 & 0.391 & \textbf{0.384} & 0.401 & 0.429 & {\ul 0.388} \\
\multirow{-3}{*}{MTS} & D.Conv & 0.012 & {\ul 0.010} & {\ul 0.010} & \textbf{0.008} & 0.012 & \textbf{0.008} \\ \midrule
 & LSTM & {\ul 0.038} & 0.039 & \textbf{0.037} & \textbf{0.037} & 0.041 & \textbf{0.037} \\
 & TsTransformer & 0.487 & 0.501 & \textbf{0.476} & 0.490 & {\ul 0.483} & 0.487 \\
\multirow{-3}{*}{UTS} & D.Conv & {\ul 0.012} & \textbf{0.011} & 0.013 & 0.015 & 0.013 & 0.020 \\ \midrule
 & \multicolumn{7}{c}{\textbf{Accuracy (Early Stopping)}} \\ \midrule
 & LSTM & 0.679 & \textbf{0.727} & 0.683 & {\ul 0.724} & 0.716 & 0.708 \\
 & TsTransformer & 0.685 & 0.700 & \textbf{0.707\dag} & 0.689 & {\ul 0.695} & 0.693 \\
\multirow{-3}{*}{MTS} & D.Conv & {\ul 0.728} & 0.716 & 0.723 & 0.711 & \textbf{0.751\dag} & {\ul 0.728} \\ \midrule
 & LSTM & \textbf{0.704} & 0.696 & 0.692 & {\ul 0.700} & 0.683$\downarrow$ & 0.692 \\
 & TsTransformer & {\ul 0.665} & 0.660 & 0.662 & 0.661 & 0.653 & \textbf{0.668} \\
\multirow{-3}{*}{UTS} & D.Conv & 0.724 & 0.721 & \textbf{0.743\dag} & 0.732 & {\ul 0.741} & 0.726 \\ \midrule
 & \multicolumn{7}{c}{\textbf{Accuracy (Last Epoch)}} \\ \midrule
 & LSTM & 0.701 & 0.711 & 0.714 & 0.716 & \textbf{0.722} & {\ul 0.719} \\
 & TsTransformer & 0.680 & 0.672 & \textbf{0.696\dag} & {\ul 0.685} & 0.680 & 0.681 \\
\multirow{-3}{*}{MTS} & D.Conv & 0.730 & 0.728 & {\ul 0.732} & 0.724 & \textbf{0.742} & 0.723 \\ \midrule
 & LSTM & \textbf{0.698} & 0.688 & 0.683 & 0.683 & 0.677 & {\ul 0.689} \\
 & TsTransformer & \textbf{0.660} & 0.650 & \textbf{0.660} & {\ul 0.656} & 0.648$\downarrow$ & 0.648 \\
\multirow{-3}{*}{UTS} & D.Conv & 0.739 & 0.740 & {\ul 0.746} & 0.745 & \textbf{0.754} & 0.738 \\ \midrule
 & \multicolumn{7}{c}{\textbf{Accuracy (Max)}} \\ \midrule
 & LSTM & 0.754 & 0.764 & 0.769 & {\ul 0.772} & 0.770 & \textbf{0.775\dag} \\
 & TsTransformer & 0.744 & 0.739 & {\ul 0.750} & 0.746 & 0.748 & \textbf{0.753} \\
\multirow{-3}{*}{MTS} & D.Conv & 0.779 & 0.772 & {\ul 0.783} & 0.779 & \textbf{0.790} & 0.775 \\ \midrule
 & LSTM & {\ul 0.770} & 0.765 & 0.763 & 0.764 & 0.761$\downarrow$ & \textbf{0.772} \\
 & TsTransformer & {\ul 0.727} & 0.722 & \textbf{0.730} & 0.725 & 0.721$\downarrow$ & 0.726 \\
\multirow{-3}{*}{UTS} & D.Conv & 0.794 & 0.796 & \textbf{0.804\dag} & 0.799\dag & {\ul 0.800} & 0.790 \\ \midrule
 & \multicolumn{7}{c}{\textbf{Accuracy (Epoch 1)}} \\ \midrule
 & LSTM & 0.355 & 0.376 & \textbf{0.456\dag} & 0.395 & {\ul 0.444\dag} & 0.435\dag \\
 & TsTransformer & {\ul 0.411} & 0.386 & \textbf{0.414} & 0.361$\downarrow$ & 0.389 & 0.397 \\
\multirow{-3}{*}{MTS} & D.Conv & 0.441 & 0.436 & {\ul 0.470} & 0.468\dag & 0.461 & \textbf{0.476} \\ \midrule
 & LSTM & 0.338 & 0.340 & \textbf{0.416\dag} & 0.388\dag & {\color[HTML]{333333} {\ul 0.396\dag}} & {\color[HTML]{333333} 0.385\dag} \\
 & TsTransformer & \textbf{0.399} & 0.361$\downarrow$ & {\ul 0.392} & 0.374 & 0.377 & 0.374 \\
\multirow{-3}{*}{UTS} & D.Conv & 0.421 & 0.409 & \textbf{0.474\dag} & {\ul 0.460\dag} & 0.458\dag & {\ul 0.460\dag} \\ \bottomrule
\end{tabular}

\caption{Main results of using different pre-training tasks on the UTS (125) and MTS (25) datasets. `R.Cls' refers to the pre-training task of Random-Cls. `Not.P' refers to the Pytorch (https://pytorch.org) default weight initialization scheme.
The Best results are in \textbf{bold} and the second best is \underline{underlined}.}
\label{tab:main_result}
\end{table*}

\begin{table*}[]
\centering
\setlength{\tabcolsep}{6pt}
\begin{tabular}{@{}llllllll@{}}
\toprule
 & \textbf{} & \textbf{Not.P} & \textbf{R.Cls} & \textbf{Ts2Vec} & \textbf{Ts-Tcc} & \textbf{Mvts} & \textbf{Srlt} \\ \midrule
\multicolumn{8}{c}{\textbf{Training Loss (Min)}} \\ \midrule
 & LSTM-underfit & 0.391 & 0.372\dag & 0.378 & 0.391 & {\color[HTML]{333333} 0.411$\uparrow$} & 0.387 \\
\multirow{-2}{*}{MTS} & D.Conv-underfit & 0.306 & 0.311 & 0.319 & 0.317 & 0.319 & 0.317 \\ \midrule
 & LSTM-underfit & 0.490 & 0.485 & 0.468 & 0.476 & 0.462\dag & 0.464\dag \\
\multirow{-2}{*}{UTS} & D.Conv-underfit & 0.414 & 0.396 & 0.401\dag & 0.410 & 0.404 & 0.404 \\ \midrule
\multicolumn{8}{c}{\textbf{Accuracy (Epoch 1)}} \\ \midrule
 & LSTM-underfit & 0.264 & 0.260 & 0.265 & 0.265 & 0.239 & 0.283 \\
\multirow{-2}{*}{MTS} & D.Conv-underfit & 0.258 & 0.276 & 0.271 & 0.297\dag & 0.308\dag & 0.303\dag \\ \midrule
 & LSTM-underfit & 0.312 & 0.303 & 0.320 & 0.313 & 0.312 & 0.311 \\
\multirow{-2}{*}{UTS} & D.Conv-underfit & 0.312 & 0.301 & 0.323 & 0.334 & 0.321 & 0.334 \\ \bottomrule
\end{tabular}

\caption{Training loss and epoch-$1$ accuracy of under-fitted LSTM and D.Conv.}
\label{tab:under_fit_LSTM_cnn}
\end{table*}

\paragraph{Effects on optimization}
Non-convex optimization in high-dimensional space has been a major challenge in deep learning since the overall process is affected by many factors~\cite{GoodBengCour16}, such as the initial parameters, the choice of the optimizer, the model structure, etc. Here we study whether the pre-trained parameters $\theta_{pre}$ lead to a lower training loss at convergence. Some studies suggest pre-training simplifies optimization~\cite{hao2019visualizing,neyshabur2020being}, while other work~\cite{erhan2010does} claim pre-training does not benefit the optimization procedure.

Analyzing Table~\ref{tab:main_result}, we can find that all pre-training tasks do not significantly obtain lower training loss compared to random initialization, that is, the pre-trained parameters do not enable the model to fit the data better.

We also analyse whether pre-training helps in optimization by considering two types of under-fitted models. Both types are poorly fitted (high training loss) but vary in structural complexity. The first type is the model with a complex structure, i.e., the TsTransformer (with a parameter count of around 400,000). The other type is the model with a simpler structure, including LSTM-underfit and D.Conv-underfit (number of parameters around 3,000). LSTM-underfit and D.Conv-underfit share the same structure with LSTM and D.Conv but differs in hidden dimensions and number of layers. Interestingly, observing Table~\ref{tab:under_fit_LSTM_cnn} and Table~\ref{tab:main_result}, we find that when pre-training the models with simple structures, pre-training can, in some cases, improve the model's fitting ability of the data. In contrast, provided with a poorly fit model with a large number of parameters, pre-training does not bring any benefit to the final optimization result.

\paragraph{Availability of regularization}
A more general definition of regularization is a technology aimed at improving the generalization ability of a model~\cite{tian2022comprehensive,GoodBengCour16}. One previous work states that the main role of pre-training is to regularize~\cite{erhan2010does}. To verify this belief, we chose three metrics to reflect the generalization ability of the model - including the accuracy in the early stopping condition, the accuracy of the last epoch, and the highest accuracy among all epochs.

We can see in Table~\ref{tab:main_result} that no pre-training task significantly improves the generalization ability of the model on both the UTS and the MTS datasets. Relatively speaking, Ts2Vec is the most effective pre-training scheme for improving generalization, but its effectiveness is only limited to some datasets and model structures, while the Mvts pre-training task even significantly degrades the generalization performance in some settings --- e.g., it degrades the performance of LSTM on the UTS dataset in the early stopping condition. Although Ts2Vec is effective in some cases, we conclude that most of the pre-training tasks for time series fail to improve the generalization ability of the model, i.e., the regularization effect of pre-training is not significant.

\begin{figure*}
    \centering
    \includegraphics[width=0.325\textwidth]{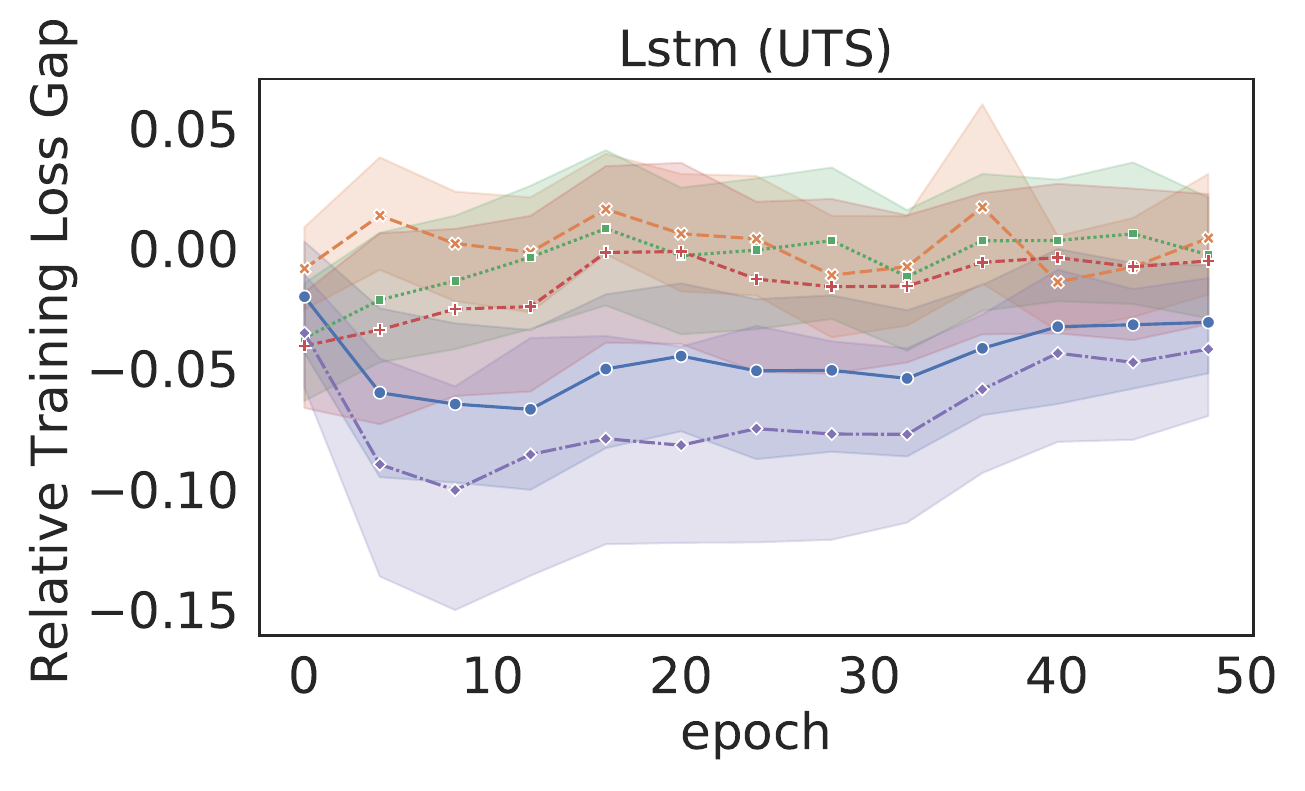}
    \includegraphics[width=0.325\textwidth]{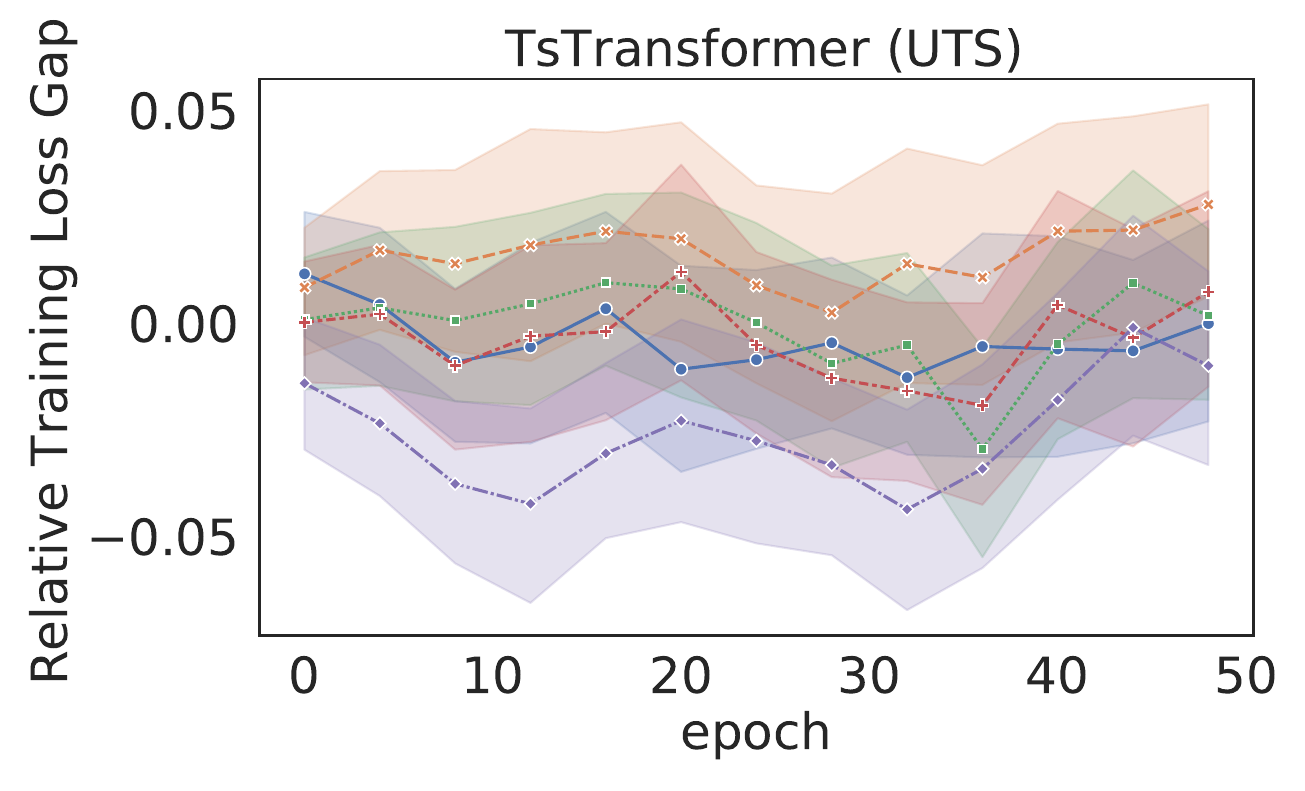}
    \includegraphics[width=0.325\textwidth]{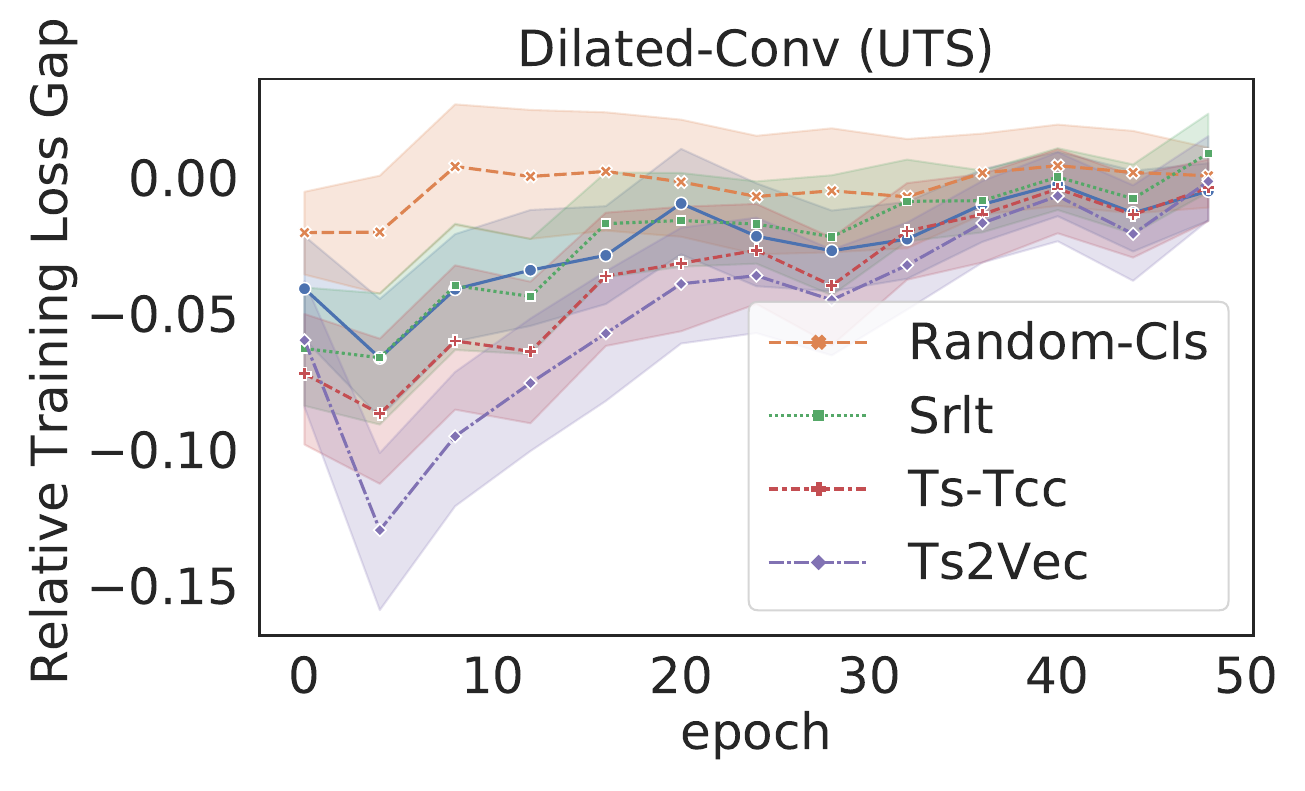}
    \includegraphics[width=0.325\textwidth]{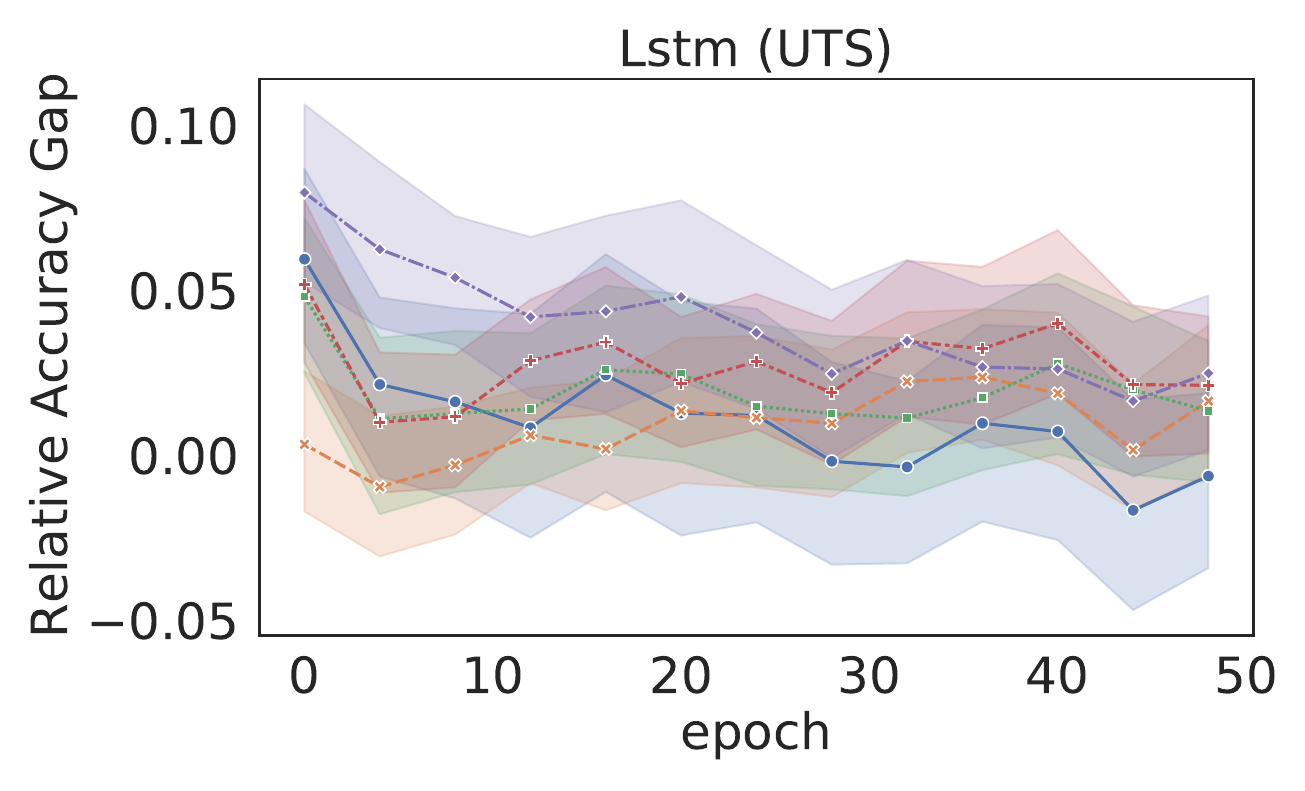}
    \includegraphics[width=0.325\textwidth]{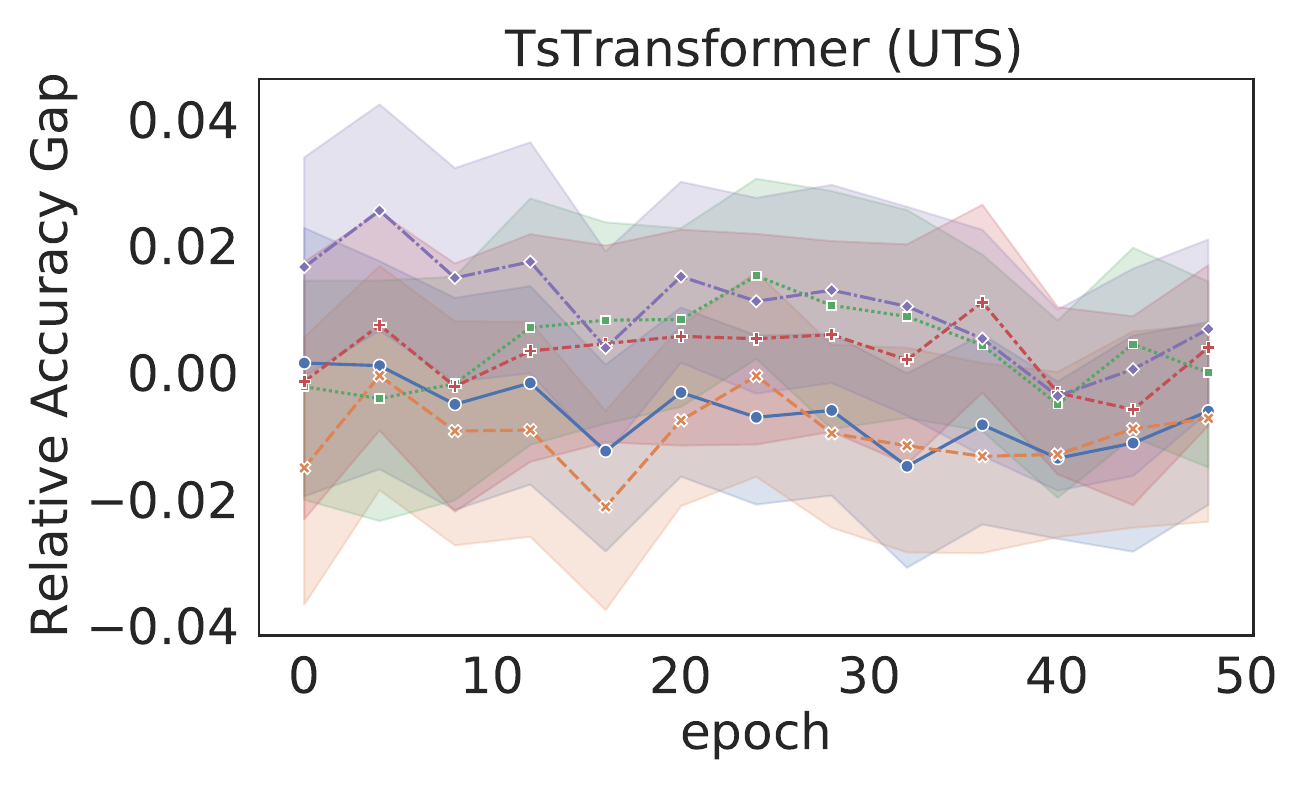}
    \includegraphics[width=0.325\textwidth]{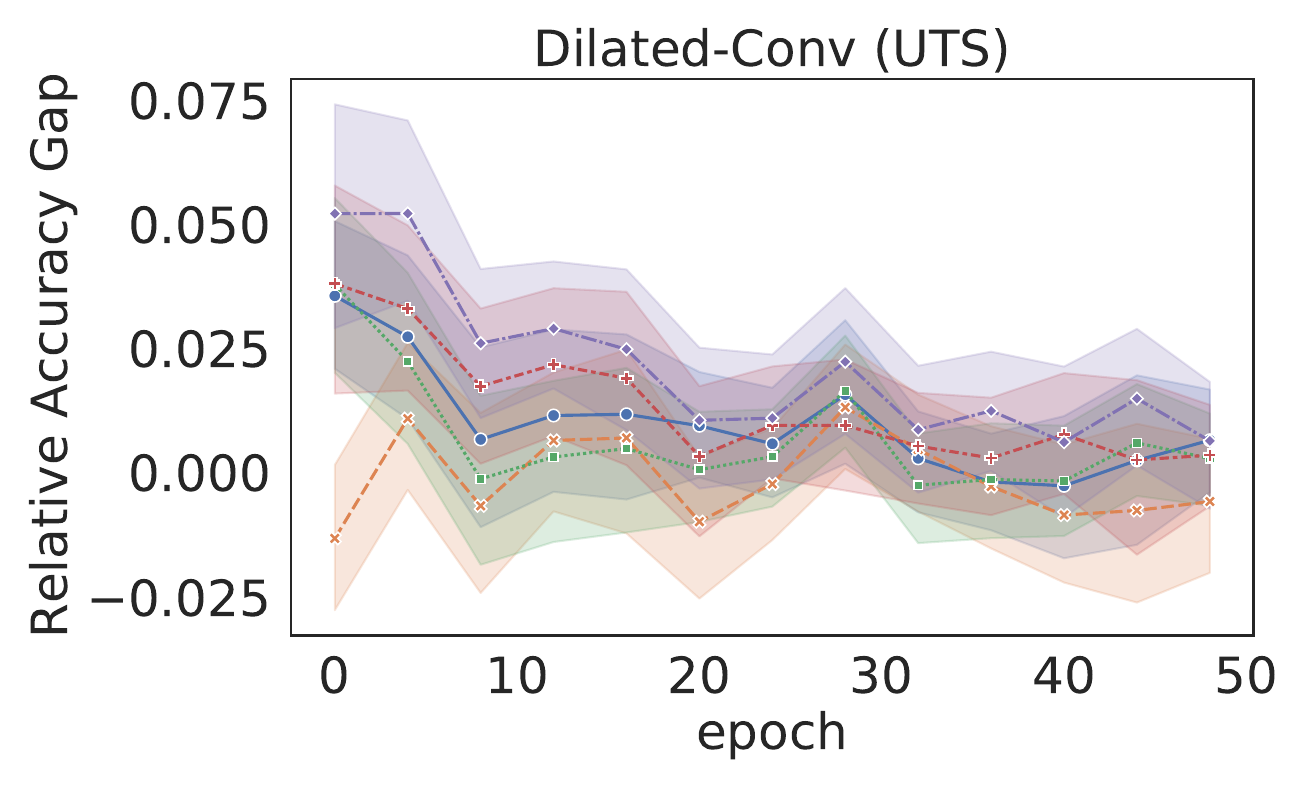}
    \caption{Gaps between each unsupervised pre-training task and random initialization in terms of convergence speed and generalization. Only the results on the UTS datasets are shown (the trend is similar on the MTS datasets). Each point is averaged across all datasets, with the length of the error bars being a $95\%$ confidence interval for the mean. The \emph{Top row} presents the decrease in training loss of the pre-trained model compared to the randomly initialized model. The \emph{Bottom row} depicts the increase in test-set accuracy.}
    \label{fig:convergence_speed}
\end{figure*}

\begin{table*}[tbh]
\centering
\setlength{\tabcolsep}{5.5pt}
\begin{tabular}{@{}lllll@{}}
\toprule
 & \textbf{Ts2vec} & \textbf{Ts-Tcc} & \textbf{Mvts} & \textbf{Srlt} \\ \midrule
\multicolumn{5}{c}{\textbf{Training Loss (Min)}} \\ \midrule
LSTM & 0.010/0.006 & 0.012/0.007 & \textbf{0.012/0.008\dag} & 0.013/0.009 \\
LSTM-underfit & 0.184/0.176 & 0.193/0.181 & \textbf{0.194/0.180\dag} & \textbf{0.193/0.179\dag} \\
TsTransformer & 0.302/0.282 & 0.301/0.287 & 0.302/0.288 & 0.303/0.296 \\
D.Conv & 0.038/0.024 & 0.046/0.025 & 0.045/0.026 & \textbf{0.056/0.021\dag} \\
D.Conv-underfit & \textbf{0.239/0.179\dag} & \textbf{0.250/0.201\dag} & \textbf{0.247/0.197\dag} & \textbf{0.246/0.202\dag} \\ \midrule
\multicolumn{5}{c}{\textbf{Accuracy (Early Stopping)}} \\ \midrule
LSTM & 0.726/0.769 & 0.737/0.787 & 0.741/0.778 & 0.736/0.780 \\
LSTM-underfit & 0.712/0.724 & 0.721/0.737 & 0.732/0.741 & 0.736/0.739 \\
TsTransformer & 0.756/0.756 & 0.745/0.745 & 0.743/0.748 & 0.746/0.742 \\
D.Conv & 0.793/0.800 & 0.786/0.807 & 0.790/0.815 & \textbf{0.783/0.819\dag} \\
D.Conv-underfit & 0.736/0.755 & 0.725/0.762 & 0.722/0.761 & 0.728/0.759 \\ \midrule
\multicolumn{5}{c}{\textbf{Accuracy (Epoch1)}} \\ \midrule
LSTM & 0.513/0.557 & 0.487/0.512 & \textbf{0.473/0.508\dag} & \textbf{0.459/0.500\dag} \\
LSTM-underfit & 0.391/0.394 & 0.378/0.376 & 0.381/0.385 & 0.377/0.383 \\
TsTransformer & 0.433/0.453 & 0.414/0.437 & 0.421/0.436 & 0.421/0.427 \\
D.Conv & 0.525/0.567 & \textbf{0.506/0.569\dag} & \textbf{0.505/0.565\dag} & \textbf{0.502/0.569\dag} \\
D.Conv-underfit & 0.397/0.422 & 0.408/0.413 & 0.406/0.417 & 0.405/0.414 \\ \bottomrule
\end{tabular}
\caption{The values before and after the slash(/) are the dataset-averaged (21 in total) accuracy of using original pre-training datasets and of using the expanded ones. The symbol $\dag$ indicates a significant advantage of using extra data.}
\label{tab:extra_data}
\end{table*}


\paragraph{Effects on convergence speed}

As we can see in Figure~\ref{fig:convergence_speed}, except when applied to TsTransformer, all pre-training schemes improve accuracy in the first few epochs and also reduce the training loss. In addition, observing the accuracy of the first epoch in Table~\ref{tab:main_result}, most of the pre-training tasks significantly boost LSTM and D.Conv at the beginning of fine-tuning. In contrast, the pre-training for TsTransformer does not show similar advantages. Since TsTransformer is under-fitted (high training loss) on most datasets, we conjecture that pre-training cannot speed up the convergence of the model when it is under-fitted. The results in Table~\ref{tab:under_fit_LSTM_cnn} largely validate our conjecture. Only a small number of cases of under-fitted D.Conv are accelerated by pre-training (using Ts-Tcc, Mvts, and srlt pre-training schemes). We summarize as follows: when the model fitting ability is sufficient, pre-training mostly improves the convergence speed and entails fast generalization, a conclusion similar to previous observations in the CV domain~\cite{he2019rethinking}, while when the model is under-fitted, pre-training usually fails to improve the convergence rate.

\paragraph{Effects of extra pre-training data}

We experiment with adding more pre-training data on three data types---\emph{Food}, \emph{ECG} and \emph{Image} (all UTS). We specifically select these three types because they contain datasets that are relatively similar to each other, which does not violate the experimental setup of in-domain pre-training. \emph{Food}, \emph{ECG} and \emph{Image} data type contains $6$, $7$ and $32$ datasets respectively. On average, the amount of pre-training data is increased by $31.41$, $27.5$, and $37.8$ folds. We illustrate the expansion process with an example. Suppose the target dataset, Beef, falls under the \emph{Food} type, we add samples of other datasets (containing both training and test sets) of the \emph{Food} type to $\mathcal{D}_{pre}$.

In Table 4 we show the impact of adding extra pre-training data. Two key points can be summarized. First, in the majority of cases, increasing the pre-training data by several times does not markedly improve the generalization ability of the model, which contradicts the previous work~\cite{paine2014analysis} that claims unsupervised pre-training helps when the ratio of unsupervised to supervised samples is high. Second, except for Ts2vec, adding more pre-training data reinforces the advantages stemming from pre-training. For instance, pre-training on the original dataset speed up the convergence of D.Conv in the early stage (Table~\ref{tab:main_result}). With additional pre-training data, the convergence speed is further accelerated. 

\subsection{Which factor, the model structure or the pre-training task, is more critical for pre-training to work on specific dataset?}

\begin{figure*}
    \centering
    \includegraphics[width=0.325\textwidth]{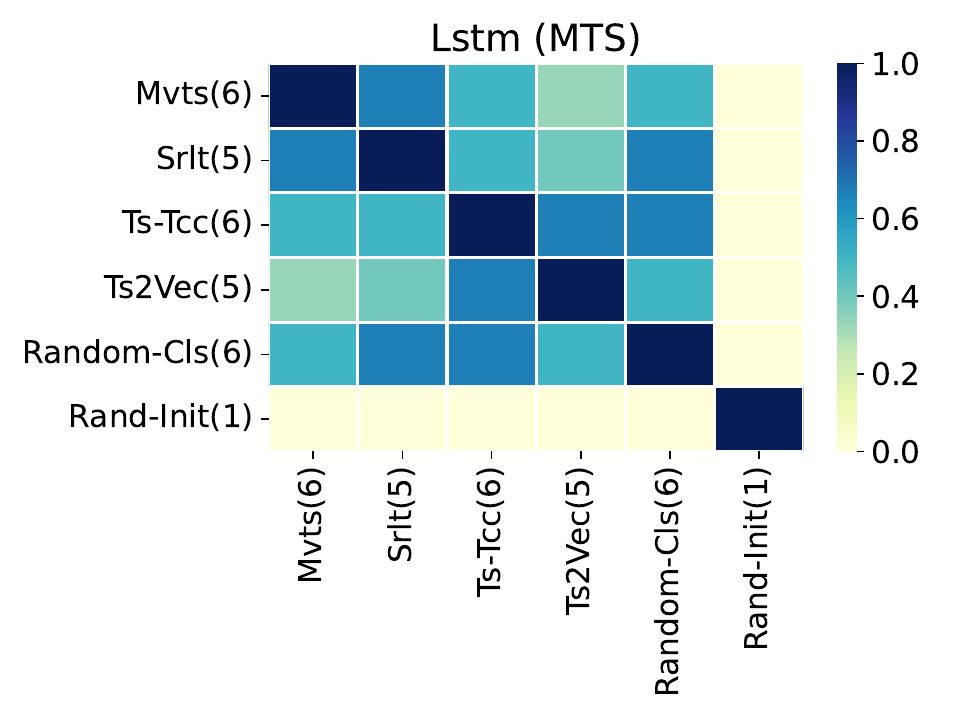}
    \includegraphics[width=0.325\textwidth]{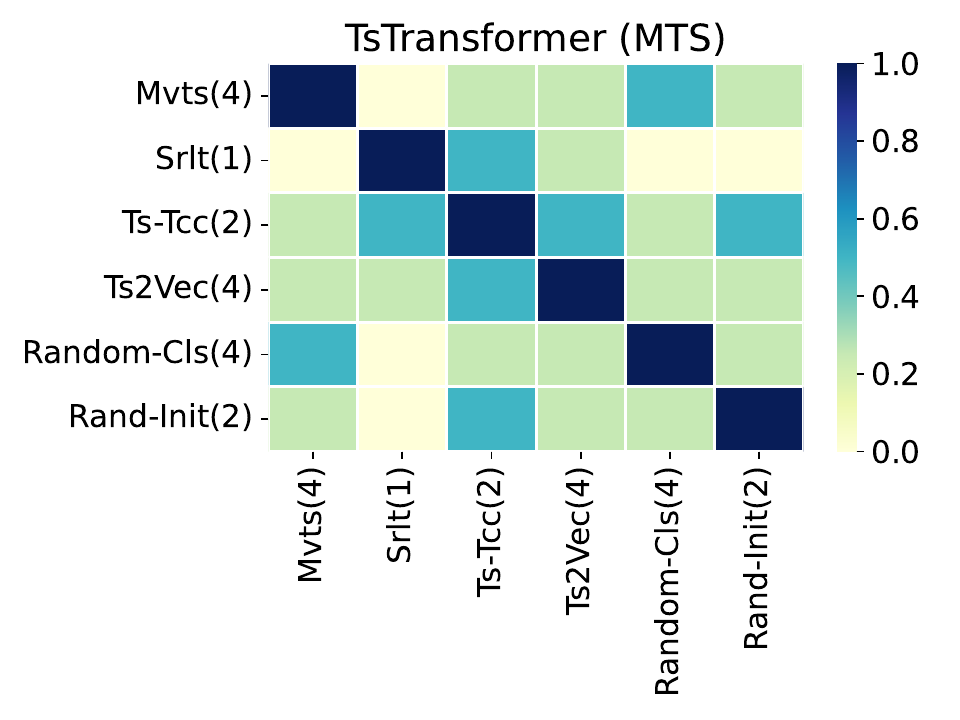}
    \includegraphics[width=0.325\textwidth]{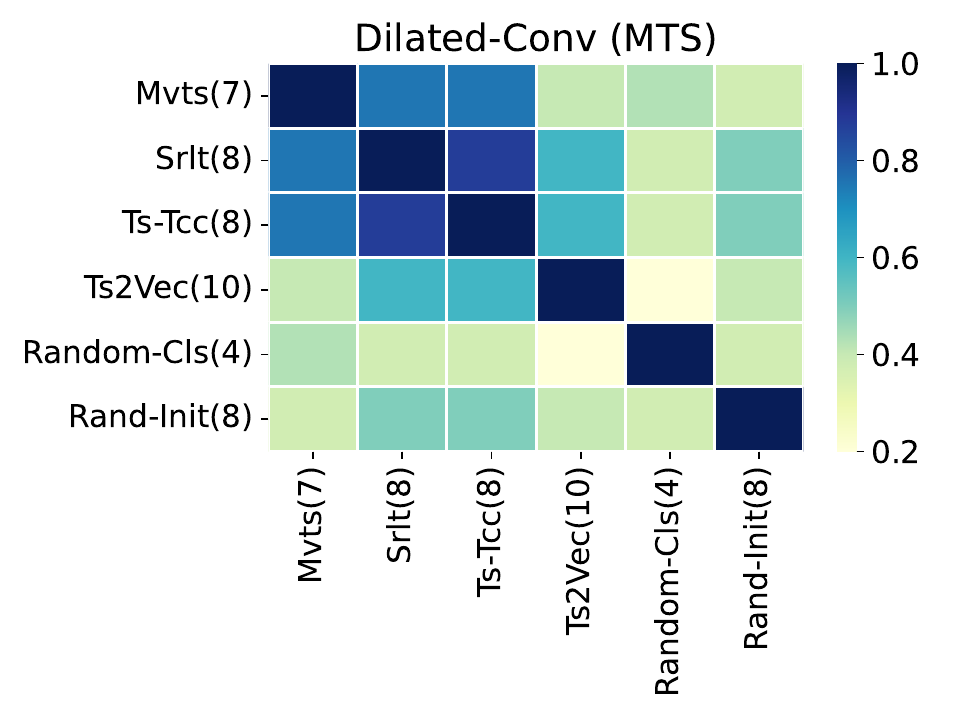}
    \includegraphics[width=0.325\textwidth]{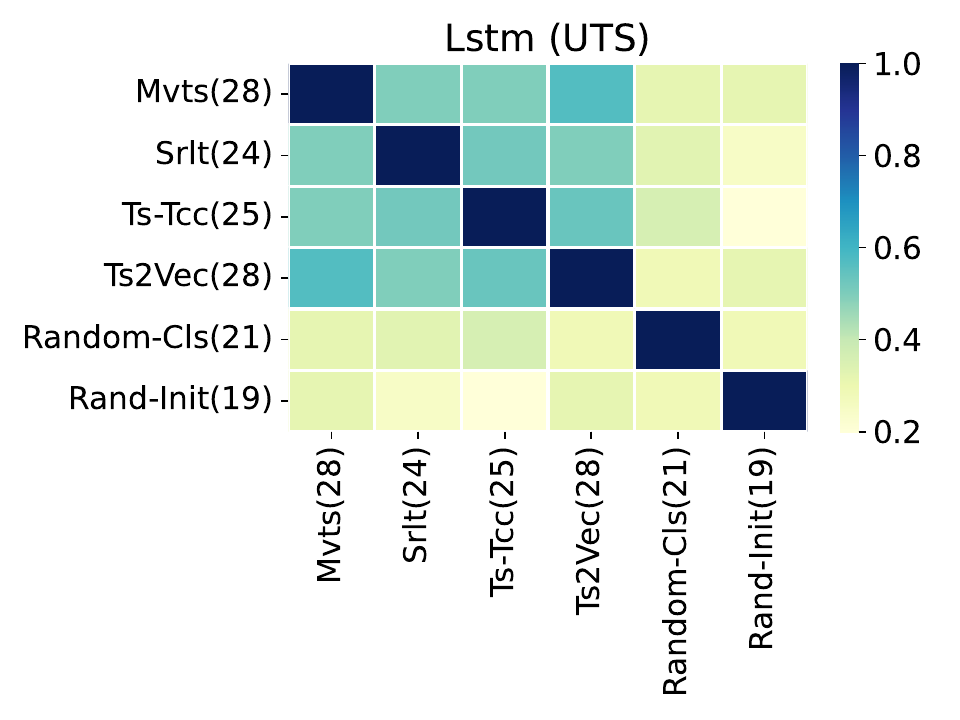}
    \includegraphics[width=0.325\textwidth]{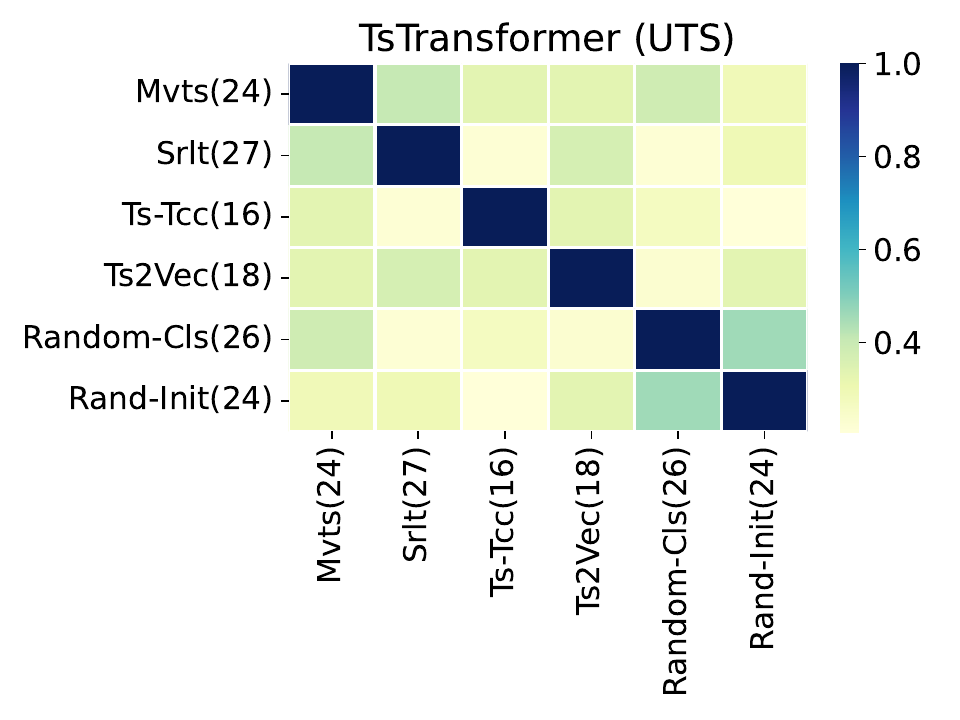}
    \includegraphics[width=0.325\textwidth]{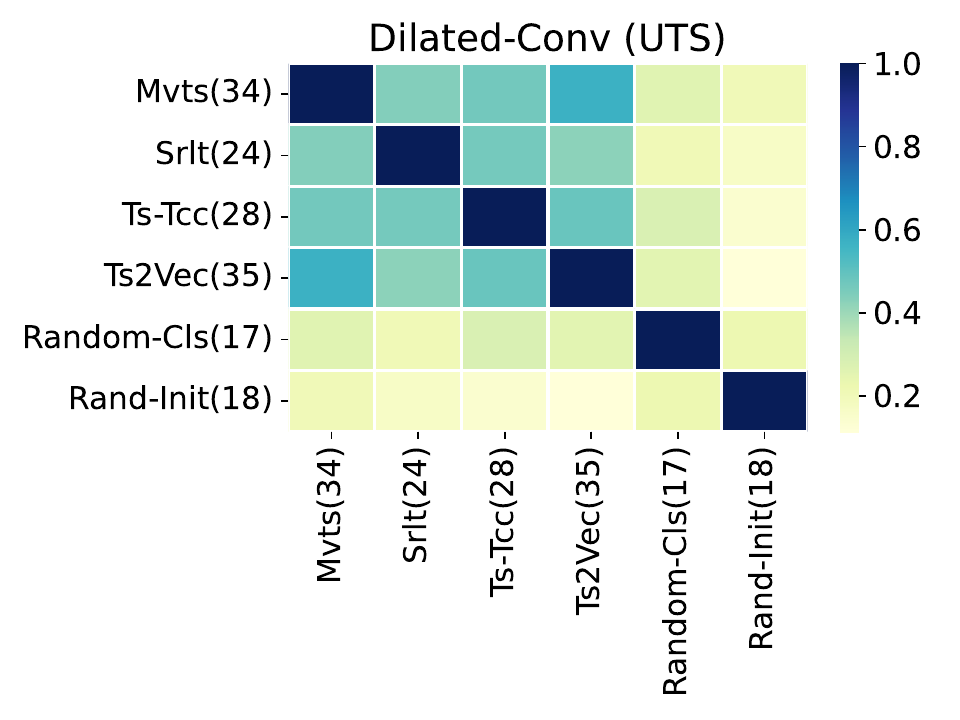}
    \caption{
    We present the intersection ratio $\varphi_{ij}$ of advantageous dataset between different pre-training tasks with the fixed model structure. The shades of color present the values of $\varphi_{ij}$. The value in parentheses represents the size of the advantageous dataset $\mathcal{A}_{pt}$. \emph{Rand-init} in this figure and \emph{R.Init} in Table~\ref{tab:best_dataset_overlap} correspond to sets of randomly initialized parameters that do not contribute to $acc_{max}$.
    }
    \label{fig:best_dataset_overlap}
\end{figure*}
Since the training of neural networks is strongly influenced by the initialized parameters~\cite{summers2021nondeterminism}, sometimes the pre-trained model outperforms the randomly initialized one by a large margin most likely because the initialized position leads to bad local minima. Thus we need a stronger baseline for locating true advantageous datasets of the pre-training task. To this end, for each dataset, we additionally train with random initialization four more times for each model structure and select the highest accuracy among the five sets of test results as the new baseline, noted as $acc_{max}$. Given a pre-training task $\mathcal{PT}$, we define its advantageous dataset set $\mathcal{A}_{pt}$ of a determined model structure as follows: for any dataset $\mathcal{D}$, if the pre-trained accuracy $acc_{pt}$ achieves a relative improvement of more than 15\% compared to $acc_{max}$, then $\mathcal{D} \in \mathcal{A}_{pt}$. We define $\varphi_{ij}=|\mathcal{A}_{pt_i} \cap \mathcal{A}_{pt_j}| /max(|\mathcal{A}_{pt_i}|, |\mathcal{A}_{pt_j}|)$ as the intersection ratio between two sets of advantageous dataset $\mathcal{A}_{pt_i}$ and $\mathcal{A}_{pt_j}$, derived from two different pre-training tasks---$\mathcal{PT}_i$ and $\mathcal{PT}_j$. We also define their intersection set as $\Omega_{ij}=\mathcal{A}_{pt_i} \cap \mathcal{A}_{pt_j}$.
We analogously define the model-wise intersection ratio $\omega_{ij}$ by substituting the pre-training task for the model structure, where $i$ and $j$ indicate different model structures.

In Figure~\ref{fig:best_dataset_overlap}, given the same model structure, except for TsTransformer, there is a high intersection ratio $\varphi_{ij}$ between each pair of the pre-training tasks (random-cls and rand-init excluded), mostly above $0.4$ and in some cases up to $0.8$ or more. More interestingly, according to statistics, such overlap is not due to the close vicinity between the pre-trained parameters. In fact, for some datasets within $\Omega_{ij}$, the $\ell_2$ distances between parameter $\theta_i$ and parameter $\theta_j$ are even further apart than their average distance across all datasets. We also calculate the common subset ratio $\phi$ among all pre-training tasks (random-cls and rand-init excluded), which is formally defined as
\begin{equation}
    \phi = |\bigcap^n_{i=1}{A}_{pt_i}|/|\bigcup^n_{i=1}{A}_{pt_i}|,
\end{equation}
where $n$ is the number of pre-training tasks. On the UTS dataset, the common subset ratio $\phi$ of LSTM, D.Conv, and TsTransformers are $0.1186$, $0.06$, and $0.1568$ respectively, which is relatively low compared with pair-wise intersection ratio $\varphi$. This result shows that the model is not the only determinant and that different pre-training tasks can bring out their advantages on different datasets.
\begin{table*}[]
\centering
\setlength{\tabcolsep}{6pt}
\begin{tabular}{@{}lllllll@{}}
\toprule
 & \textbf{Ts2Vec} & \textbf{Ts-Tcc} & \textbf{Mvts} & \textbf{Srlt} & \textbf{R.Cls} & \textbf{R.Init} \\ \midrule
MTS & \textbf{.233$\pm$.047} & .125$\pm$.177 & {\ul .151$\pm$.117} & .125$\pm$.102 & .139$\pm$.104 & .042$\pm$.059 \\ \midrule
UTS & .252$\pm$.086 & .193$\pm$.099 & \textbf{.301$\pm$.085} & {\ul .272$\pm$.046} & .198$\pm$.070 & .136$\pm$.037 \\ \bottomrule
\end{tabular}
\caption{We present overlap ratio $\omega_{ij}$ of advantageous datasets between different model structures when the pre-training task is fixed. Here we interpret $i$ and $j$ as two different model structures.}
\label{tab:best_dataset_overlap}
\end{table*}

Analyzed from another perspective, Table~\ref{tab:best_dataset_overlap} shows that the model-wise overlap $\omega_{ij}$ is relatively small when we fix the pre-training task. Nevertheless, we still find that Mvts and Ts2Vec methods significantly outperform two baselines --- R.Cls and R.Init, demonstrating a potential of luring the model to perform better on certain datasets regardless of its structure. 

Observing Table~\ref{tab:best_dataset_overlap} and Figure~\ref{fig:best_dataset_overlap}, if the question is how to make pre-training benefit fine-tuning on a certain dataset, it is more important to consider the fit of the model structure rather than designing a specific pre-training task, because the model structure has a much greater impact than the pre-training task for squeezing the potential of unsupervised pre-training on a given dataset.


\subsection{Correlation Factors}

\begin{table*}[tbh]
\centering
\setlength{\tabcolsep}{3.2pt}
\begin{tabular}{@{}cccccccccc@{}}
\toprule
 &  & \textbf{Seq.L} & \textbf{Pre.S} & \textbf{$\ell_2$-norm} & \textbf{$\ell_2$-path} & \textbf{Sharp.} & \textbf{P.T} & \textbf{P.V} & \textbf{Dis.} \\ \midrule
\multicolumn{10}{c}{\textbf{Mvts}} \\ \midrule
\multirow{3}{*}{UTS} & Lstm & 0.045 & -0.020 & 0.017 & 0.017 & -0.095 & 0.162 & \textbf{0.198\dag} & 0.027 \\
 & Transf. & -0.036 & 0.137 & -0.078 & -0.086 & 0.172 & -0.069 & -0.080 & -0.027 \\
 & D.CNN & 0.0270 & -0.147 & 0.093 & 0.094 & \textbf{0.229\dag} & 0.109 & 0.115 & -0.130 \\ \midrule
\multicolumn{10}{c}{\textbf{Srlt}} \\ \midrule
\multirow{3}{*}{UTS} & Lstm & -0.037 & 0.058 & 0.040 & 0.037 & 0.174 & -0.060 & -0.043 & 0.045 \\
 & Transf. & -0.002 & -0.035 & 0.049 & 0.047 & -0.133 & 0.001 & -0.106 & -0.102 \\
 & D.CNN & -0.127 & -0.004 & -0.040 & -0.042 & 0.112 & -0.009 & -0.071 & -0.050 \\ \midrule
\multicolumn{10}{c}{\textbf{Ts-Tcc}} \\ \midrule
\multirow{3}{*}{UTS} & Lstm & -0.070 & 0.080 & 0.182 & \textbf{0.218\dag} & 0.085 & 0.002 & -0.052 & 0.009 \\
 & Transf. & 0.0381 & -0.061 & -0.022 & -0.021 & -0.094 & 0.176 & -0.047 & 0.016 \\
 & D.CNN & -0.037 & -0.067 & 0.049 & 0.051 & 0.131 & -0.109 & -0.066 & -0.046 \\ \midrule
\multicolumn{10}{c}{\textbf{Ts2Vec}} \\ \midrule
\multirow{3}{*}{UTS} & Lstm & -0.211 & 0.053 & 0.039 & 0.041 & 0.169 & -0.024 & -0.024 & 0.040 \\
 & Transf. & -0.010 & 0.078 & -0.002 & 0.001 & 0.039 & -0.005 & 0.033 & -0.066 \\
 & D.CNN & 0.057 & -0.225 & -0.104 & -0.109 & 0.046 & 0.087 & 0.102 & \textbf{-0.195\dag} \\ \midrule
\end{tabular}

\caption{Spearman correlation coefficient between different factors and accuracies (early-stopping) of the test set. The results in this table are derived from the UTS datasets only, But the situation is similar on the MTS datasets. \textbf{Seq.L} is sequence length. \textbf{Pre.S} is the size of the pre-training set. \textbf{Sharp.} means sharpness. \textbf{P.V} represents the convergence state of the model on the pre-training set and is equal to the lowest training loss divided by the training loss of the first epoch. \textbf{P.V} is defined similarly on the validation set. \textbf{Dis.} is the $\ell_2$ distance between the pre-trained parameters and its initial point.}
\label{tab:correlation}
\end{table*}

From Table~\ref{tab:correlation}, we can see that there is not a single factor that is significantly associated with the effectiveness of pre-training. There are some sporadic significant associations, but they are only observed on a small number of model structures, which do not provide much of a reference. This result is rather disappointing, and we have yet to find an indicator to guide us on whether to trust the pre-trained parameters or to decide whether to pre-train on certain datasets.

\section{Conclusion and Future Work}
This paper focuses on the study of whether unsupervised pre-training is beneficial for fine-tuning on the Time Series Classification task, making an empirical contribution to the study of when and how unsupervised pre-training helps fine-tuning. We conclude that pre-training does not significantly enhance the generalization performance but can improve the convergence speed when the model fits the data well. Also, it can improve the optimization process for simple models with a small number of parameters. Increasing the amount of pre-training data does not benefit generalization, but amplifies the existing advantages of pre-training, such as fast convergence. When dealing with a new time series dataset and aiming to enhance the model using pre-training, it's more important to focus on creating an appropriate model architecture than on creating the pre-training task itself.

The experimental procedures and conclusions presented in this paper are primarily grounded in low-resource settings, utilizing medium-sized models and limited pre-training data. This approach is particularly relevant to the current discourse surrounding artificial intelligence, as the development of AI with lower resources has become a pressing concern for various fields, including natural language processing~\cite{rotman2019deep}, healthcare~\cite{wahl2018artificial,pokaprakarn2022ai}, edge computing~\cite{merenda2020edge}, and green AI~\cite{schwartz2020green}. To extend the scope of our work, we encourage future studies to explore larger models and more data in this domain, as well as to do new research on catastrophic forgetting~\cite{mehta2021empirical}, intrinsic dimension~\cite{aghajanyan2021intrinsic}, and other aspects not covered in this paper.



\bibliography{main}
\end{document}